# Passive Control Architecture for Virtual Humans

Antoine Rennuit, Alain Micaelli, Xavier Merlhiot, Claude Andriot
*CEA\LIST*
*Fontenay-aux-Roses, France*

{rennuita, micaellia, merlhiotx, andriotc}@zoe.cea.fr

Antoine Rennuit, François Guillaume, Nicolas Chevassus
*EADS\CCR*
*Suresnes, France*

{antoine.rennuit, francois.guillaume, nicolas.chevassus}@eads.net

Antoine Rennuit, Damien Chablat, Patrick Chedmail
*IRCCyN*
*Nantes, France*

{Antoine.Rennuit, Damien.Chablat, Patrick.Chedmail}@irccyn.ec-nantes.fr

*Abstract* – **In the present paper, we introduce a new control architecture aimed at driving virtual humans in interaction with virtual environments, by motion capture. It brings decoupling of functionalities, and also of stability thanks to passivity.**
**We show projections can break passivity, and thus must be used carefully.**
**Our control scheme enables task space and internal control, contact, and joint limits management. Thanks to passivity, it can be easily extended.**
**Besides, we introduce a new tool as for manikin's control, which makes it able to build passive projections, so as to guide the virtual manikin when sharp movements are needed.**

*Index Terms* – *Virtual humans, Contact, Passive projections, Motion capture.*

## I. INTRODUCTION

The control of Virtual Humans (VH) is still very limited. The computer graphics industry knows how to produce outstanding images, but at the price of a long animation step.

Our main purpose is the virtual manikin for engineering; we focus our work on interactive animation that makes it able to drive an avatar in Real Time (RT) through a motion capture device, as seen on Fig. 1. The main problem being the dimension of motion capture signals is much shorter than the system's dimension.

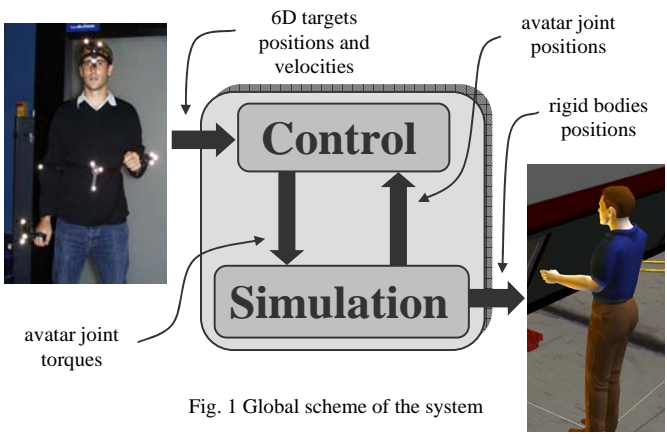

Fig. 1 Global scheme of the system

One of the main issues we will have to deal with is the problem of *conflicting retargeting*. This problem is encountered when one want a movement acquired from a given actor in the motion capture, to be applied on a virtual actor that has a completely different morphology. Let's give an example. The hands and feet of a giant actor are tracked, and we want their motion to be applied on a dwarf's hands and feet: there will be a conflicting situation if the giant raises his hands *too high*, because the dwarf will not be able to reach all targets at the same time.

A first answer to the retargeting problem was brought by Gleicher in [5]. His work enables to retarget the motion of a puppet onto another differently morphologied puppet while preserving constraints (such as hands, and feet positions), and optimizing criteria defined such that they refine the movement. This work is based on space-time optimization that is the final motion is optimized on a whole sequence at once given an initial motion that must be known before computing. This point prevents us from using this technique. Moreover, his technique does not consider forces, and interaction.

Popovic and Witkin [6] describe a technique allowing to retarget motion, it is also based on space-time optimization; they include the dynamics equation as a constraint in their optimization scheme, to preserve the physical nature of the motion. The optimization problem is much heavier, thus they reduce the problem to a small number of degrees of freedom, excessively simplifying the problem. It also suffers from the space-time problem stated above.

Both Gleicher, and Popovic's methods can handle retargeting problems only when the retargeting is not conflicting.

Baerlocher and Boulic [7], noticed that when conflicting situations occur, some constraints are more important than others. Let's take the example of the giant and the dwarf again. In conflicting situations, we would better satisfy the constraints of the feet, than the ones on the hands, because to obtain a feasible realistic movement, a character keeps feet on the ground. Thus their work was dedicated to the introduction of priorities on constraints. This method is not based on space-time optimization anymore, but on an Inverse Kinematics (IK) technique. The main problem with IK is its lack of support for forces.

This problem is solved thanks to Sentis, and Khatib [8], who propose tasks prioritization within the dynamical context. Their approach is to decouple the Lagrange dynamic equation thanks to projections of this equation into the kernel of the Jacobian matrices of higher priority tasks. Although they do not tackle with contact, and interaction with environment, their approach is interesting to solve our problem.

In our framework, we would like to be able to drive, in a "realistic way", a manikin, doing sharp tasks such as the ones that can be done by a worker, e.g. screwing a screw, sawing, drilling a hole, or nailing down a nail… These tasks all need sharp movements of the worker in the real world, and need the same accuracy in a virtual world.

Thus to be able to be "realistic", we will have to emulate the real world's physical laws, and human specificities: interaction with environment, non-penetration of objects, joint limits enforcement, human-like motion of the avatar…

The main point in the features we want to be implemented is *interaction with environment*, because it drives the choice of the model we are to put into action. If we want to be able to interact in a natural way, we have to implement natural behaviors. That is interaction must be done through forces. This means kinematical approaches are not adequate: we must use the equations of dynamics.

This perfectly fits the general context of our work: our architecture will be coupled to a portable haptic device being developed in [12], as seen on Fig. 2.

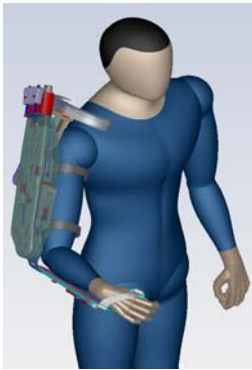

Fig. 2 Portable haptic device that will be integrated into our framework

The system we aim at controlling is complex. An effective approach in such cases is to implement a modular architecture, so as to decouple functionalities of our systems in different modules. This will be made possible thanks to a passive approach. Actually, passivity will allow us to bring the modularity a step further than the simple functional modularity. Indeed it will allow decoupling the analysis of the stability of the whole system, into the analysis of the passivity of each module.

The approaches enabling tasks prioritization such as [8] seem interesting in case of conflicting tasks. Unfortunately, they cannot be used in the context of passivity, because they use projections. As we will show, in the general case, the use of projections while optimizing other potentials is not compatible with passivity: in general projections break passivity. We make things clear in II.

This could seem very disturbing. Nevertheless, this limit only appears in case of impossibility for the avatar to reproduce the movements of the actor. In the context of engineering (at least), one does not really want to control the virtual human in the case of unfeasible movements. Indeed what we really want is knowing if the movement is feasible or not, this makes a big difference.

Knowing this, we propose other control modes, which help the manikin achieving its movements, instead of trying to control it performing movements it will never succeed in. These control modes are relevant because of a loss of information when we are immersed in simulations. Let's take the example of the worker again. When drilling its hole in the real world, in a real environment (let's say a wall), the worker is helped aligning the drill axis on the hole axis thanks to its haptic sensation. Unfortunately, this sensation cannot help anymore when performing the motion in the motion capture device, because the virtual environment which is to be drilled, has no real counterpart.

Haptic devices' approach is interesting, but requires an heavy infrastructure. Thus, if we wanted to be able to perform precision control, with a light infrastructure we would use virtual guides, based on projections. Nevertheless, as explained above, we will show that applying a projection "as is" can lead to the loss of the passive nature[1] of our controller, so we propose a way to build *physical projections*, which respect passivity, thanks to mechanical analogies.

We also implement a solution to solve for contacts. Zordan, and Hodgins [10], propose a solution that "hits and reacts", modifying control gains during the simulation. This leads to an approach that is known to be unsafe because of instabilities, and is not real-time. Schmidl, and Lin [11] use a hybrid they call geometry-driven physics that uses IK to solve for the manikin's reaction to contact, and impulse-based physics for the environment. They lose the physical nature of the simulation. The method we use does not make these concessions.

We are now about to discuss the whole control framework architecture in the passivity context. Then we will introduce our passive virtual guides, and test our solution.

II. DETAILED CONTROL SCHEME

The global control scheme we want is depicted in Fig. 1. We know motion capture positions are the inputs of a controller (which is about to be detailed), this controller drives a physical simulation (also to be detailed); which sends the updated world configuration to the output renderer.

The manikin we aim at controlling is composed of two layers: a skeleton (which can be viewed as a kinematical chain), and a rigid skin on top of it, which will be useful for collision detection (of course, once motion is calculated, it can be sent to a nice renderer which will tackle with soft deforming skin – which is out of scope).

Now we describe and motivate the approach we chose. The scheme Fig. 3, shows the whole architecture.

---

[1] **loss of the passive nature :** this point goes beyond the stability issue. Real world physical phenomenons are intrinsically passive, hence the loss of passivity also means the loss of the physical nature of the simulation [17].

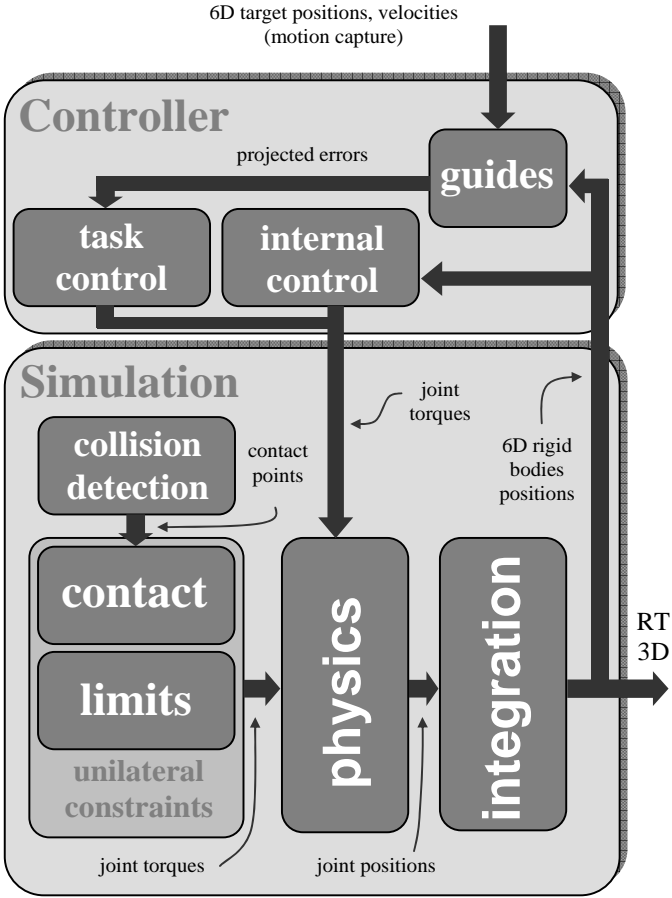

Fig. 3 Detailed global control scheme

## A. Simulation

The blocks *physics*, and *integration* of the scheme Fig. 3 should be self-motivated, as we want to emulate the physical laws of the real world. Let's just describe some choices we made.

*1) Physics*

As stated above, the simulation is to support contact, and interaction with the environment, that is forces. Hence dynamics established itself as the best choice for our model. In a first attempt, we have used a first order dynamical model[2], as stated in [3] (though simpler it highlights the problem to be solved). Integration is done through an additional joint damping term, that is

$$A(q)\ddot{q} + C(q,\dot{q})\dot{q} + G(q) + B_a\dot{q} = \Gamma, \quad (1)$$

becomes
$$B_a\dot{q} = \Gamma, \quad (2)$$

with $B_a$, the damping matrix chosen to be symmetric positive semi-definite; $q$, and $\Gamma$ are the joint parameters and torques.

Of course, these physical laws are intrinsically passive at port $<\dot{q}, \Gamma>$, the proof can be found in [1].

*2) Integration*

---

[2] **Neglect non-linear terms :** that leads to first order dynamics, or dynamics with no mass (though weight can be considered, inertial and Coriolis effects are neglected).

We use a Runge-Kutta-Munthe-Kaas scheme (see [4]). This is a Runge-Kutta method dedicated to integration on Lie Groups, that is known to be efficient. This work is left to Generalized Virtual Mechanisms (GVM) [14], a library being developed in CEA\LIST.

*B. Control*

In order to understand the relevance of the other functional blocks of the diagram, we have to wonder *what makes a human move, and have the specific motion he has*? In doing so, we found three sources of movement (or influential factors):
- The first one enables an end effector to reach a goal in task space:
    Ex: I want my left hand to reach a plug on the wall.
- The second one drives configuration, or gaits
    Ex: it makes the difference between the gait of a fashion model, and the gait of an old cowboy.
- The last one enforces physical, and biomechanical constraints:
    Ex: joint limits, non-penetration with environment, but also balance control (which will be taken into account in a forthcoming paper as in [15] )…

These influential factors can be translated straightforwardly into control idioms, being task space, and null-space control under unilateral constraints.

Bilateral and unilateral constraints will not cause any problem. But operational space, and null-space control must be studied carefully if one does not want to break passivity.

*1) External task space control*

As explained in section I, at first glance it could seem interesting to bring prioritization between external tasks. Nevertheless, we show that such prioritizations can break passivity. This urges us to use other control modes.

In [8], Sentis, and Khatib introduce dynamical decoupling of *n* external tasks. The joint torques $\Gamma$ they apply on their manikin is composed of the influence $\Gamma_i$ of *n* prioritized external tasks, such that lower priority tasks do not disturb higher ones:

$$\Gamma = \sum_{i=1}^{n} \Gamma_{i|prev(i)}, \text{ with } \Gamma_{i|prev(i)} = \Pi^T_{prev(i)}\Gamma_i, \quad (3)$$

with $\Pi^T_{prev(i)}$ projecting into higher priority tasks' null-space.

Using such projections is unsafe, we show they can break passivity:

Let's take an example with two external ports $(J_1, W_1, V_1, \Gamma_1)$, and $(J_2, W_2, V_2, \Gamma_2)$, each port being described by its Jacobian, the wrench applied, its velocity, and the torques it generates. We give task 1 the highest priority, and $\Pi_1$ is the projection allowing to enforce priorities, which is to be defined, thus we can write:

$$\begin{cases} \Gamma_1 = J_1^T W_1 \\ \Gamma_2 = \Pi_1^T J_2^T W_2 \end{cases},$$

and
$$V_1 = J_1\dot{q} = J_1 B_a^{-1}(\Gamma_1 + \Gamma_2),$$

$$= J_1 B_a^{-1} J_1^T W_1 + J_1 B_a^{-1} \Pi_1^T J_2^T W_2. \quad (4)$$

The priority appears if $\Pi_1^T$ projects into $Ker(J_1 B_a^{-1})$.

For the passivity to be ensured, we must enforce:
$$\int_0^t (W_1^T V_1 + W_2^T V_2) dt \geq -\beta^2, \text{ with } \beta \text{ real}, \forall W_1, \text{ and } W_2. \quad (5)$$

We develop $W_1^T V_1 + W_2^T V_2$:

$$W_1^T V_1 + W_2^T V_2 =$$
$$= (W_1^T J_1 + W_2^T J_2) \dot{q}$$
$$= (W_1^T J_1 + W_2^T J_2) B_a^{-1} (J_1^T W_1 + \Pi_1^T J_2^T W_2)$$
$$= (W_1^T J_1 B_a^{-1} J_1^T W_1 + W_2^T J_2 B_a^{-1} J_1^T W_1 + W_2^T J_2 B_a^{-1} \Pi_1^T J_2^T W_2)$$
$$= \left(W_1^T J_1 + \frac{1}{2} W_2^T J_2\right) B_a^{-1} \left(J_1^T W_1 + \frac{1}{2} J_2^T W_2\right)$$
$$- \frac{1}{4}(W_2^T J_2 B_a^{-1} J_2^T W_2) + (W_2^T J_2 B_a^{-1} \Pi_1^T J_2^T W_2), \quad (6)$$

There always exists $W_1$, $W_2$, and $J_2$, such that:
$$\begin{cases} J_2^T W_2 \neq 0 \\ J_1^T W_1 + \frac{1}{2} J_2^T W_2 = 0 \\ J_2 B_a^{-1} \Pi_1^T J_2^T = 0 \end{cases}, \quad (7)$$

A trivial example can be shown when $J_2 = J_1$, in this case, we only have to choose $W_1 = \frac{1}{2} W_2$.

Equation (7) implies that (6) becomes:
$$W_1^T V_1 + W_2^T V_2 = -\frac{1}{4}(W_2^T J_2 B_a^{-1} J_2^T W_2), \quad (8)$$

(8) implies $\int_0^t (W_1^T V_1 + W_2^T V_2) dt$ has no lower bound, so .

(5) cannot be enforced. Thus in the general case, projecting external interactions can break passivity.

As explained earlier, projections are useful in case of conflicting targets, in the case when the manikin cannot achieve what it is asked to do. It means that a real human with the same morphology as its virtual counterpart, could neither achieve the movement. In the case of engineering, we do not look for controlling virtual humans, in cases where they cannot achieve the target motion. We only want to know if the movement is feasible or not. Thus the proposed control is rather simple, but behaves well in case of unfeasible movements, and is able to warn in case of infeasibility.

In [13] they use a 6D Proportional Derivative (PD) operational space controller at each point to be controlled on the manikin (Fig. 4):
$$f_{ctrl} = K(x_d - x) + B_c(v_d - v). \quad (9)$$

Note that if the control points where linked to their targets position, thanks to a damped spring, the force generated would have the same shape, this makes it able to "draw" the controller, as its mechanical analogy, the damped spring.

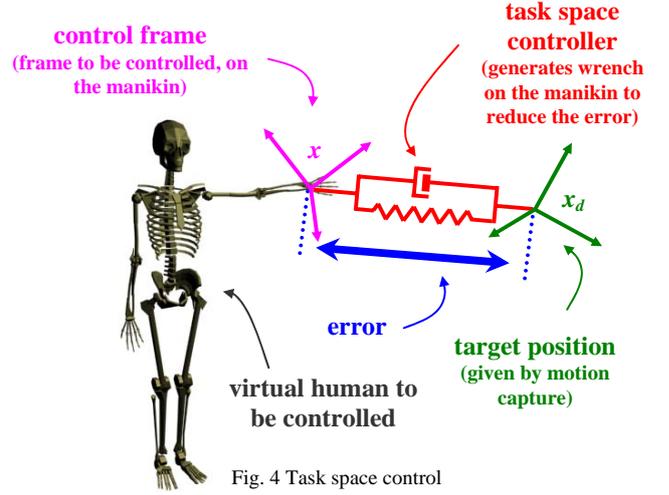

Fig. 4 Task space control

Concatenating (2), and (9), we obtain:
$$B_a \dot{q} = J^T (K(x_d - x) + B_c(v_d - v)),$$
that is: $(B_a + J^T B_c J) \dot{q} = J^T (K(x_d - x) + B_c v_d)$.

This equation admits a solution if $(B_a + J^T B_c J)$ is not singular, that is $B_a$ must be definite, or $J$ must be full rank, with $B_c$ definite, then:
$$\dot{q} = (B_a + J^T B_c J)^{-1} J^T (K(x_d - x) + B_c v_d). \quad (10)$$

*2) Constraints*

All the constraints enumerated above (joint limits, contact…) are unilateral. They can all be solved through Linear Complementary Problems (LCP) algorithms. We use GVM's unilateral constraints solver [14].

Using an approach similar to Ruspini, and Khatib [2], we express the contact problem in an LCP form, which is solved for $f$ - the contact wrench. This stage is passive so long as the dynamical equation is passive.

*3) Configurations (internal control)*

Null space control is usually solved as in [9], optimizing internal potentials (the choice of these potentials is out of scope). If we do not want this optimization to disturb external control, we must work in the null space of our external task, projecting the joint torques induced from the internal potential. Here we show this projection can break passivity.

Let's take a skeleton, an internal potential $U(q)$, to be minimized (its associated joint torques are $\Gamma_{int}$), and an external port (defined as in 1) by $(J_1, W_1, V_1, \Gamma_1)$). Projection $\Pi_1$ (to be defined) is to give priority to the external task:
$$\Gamma_{int} = -\alpha \Pi_1^T \frac{\partial U}{\partial q}^T.$$

Speed at the external port 1 is:
$$V_1 = J_1 \dot{q} = J_1 B^{-1} \Gamma = J_1 B_a^{-1} \left(J_1^T W_1 - \alpha \Pi_1^T \frac{\partial U}{\partial q}^T\right), \quad (11)$$

If one does not want the internal potential to disturb the external, we must take $\Pi_1^T$ as a projection in $Ker(J_1 B_a^{-1})$.

We must ensure this projection keeps passivity at all ports. At the external port, thanks to (11) we can write:

$$W_1^T V_1 = W_1^T J_1 B_a^{-1} J_1^T W_1 \geq 0. \quad (12)$$

So the system composed of the external task coupled to the internal one is passive at the external port, if the external port was passive before coupling.

At the internal port:

$$\dot{U} = \frac{\partial U}{\partial q} \dot{q} = \frac{\partial U}{\partial q} B_a^{-1} \left( J^T W_1 - \alpha \Pi_1^T \frac{\partial U}{\partial q}^T \right),$$

that is: $\dot{U} = \frac{\partial U}{\partial q} B_a^{-1} J_1^T W - \alpha \frac{\partial U}{\partial q} B_a^{-1} \Pi_1^T \frac{\partial U}{\partial q}^T. \quad (13)$

It's passive if $B_a^{-1} \Pi_1^T$ Symmetric Positive Definite (SPD), which we will proven when taking $\Pi_1^T$ orthogonal such as:

$$\Pi_1^T = I - (J_1 B_a^{-1})^T (J_1 B_a^{-1})^{+T}. \quad (14)$$

Then: $B_a^{-1} \Pi_1^T = \sqrt{B_a^{-1}} \Pi_1^T \sqrt{B_a^{-1}}$,

$\Pi_1^T$ being orthogonal, it is SPD, thus $B_a^{-1} \Pi_1^T$ is SPD.

Problems arise when the skeleton has more than one port. Imagine that the skeleton collides the environment at port 2, $(J_2, W_2, V_2, \Gamma_2)$. We can write (we assume $W_1 = 0$ for simplicity): $W_2^T V_2 = W_2^T J_2 B_a^{-1} \Gamma$

$$= W_2^T J_2 B_a^{-1} J_2^T W_2 - \alpha W_2 J_2 B_a^{-1} \Pi_1^T \frac{\partial U}{\partial q}^T.$$

The internal potential's influence does not disappear as in (12). Then, projecting an internal potential can break passivity.

Nevertheless, there are solutions to this problem:
- We can reduce the internal task's influence through $\alpha$ such that the system remains passive.
- We can use *self-projective* internal potentials, which are potentials such that:

$$\Pi_1^T \frac{\partial U}{\partial q}^T = \frac{\partial U}{\partial q}^T. \quad (15)$$

This notion is related to the potential's projected gradient's integrability (see [16]).

- Extended projections $\Pi_{prev(i)}^T$ can also be used, they project in all external ports' null space.

Such solutions are not yet implemented in our control; the internal dynamic is left open-loop for the time being. However, we can tune the configuration through $B_a$.

### III. PASSIVE VIRTUAL GUIDES

As stated before, we want to add the possibility to guide the movements of our virtual human (*guides* block of Fig. 3). The most intuitive way to implement such guides, is to project the error to be corrected by our operational space controller.

We showed introducing the projection matrix, could break passivity. As in telerobotics [3], we used the virtual link concept, in order to realize passive projections. Following the mechanical analogy approach, passivity is ensured.

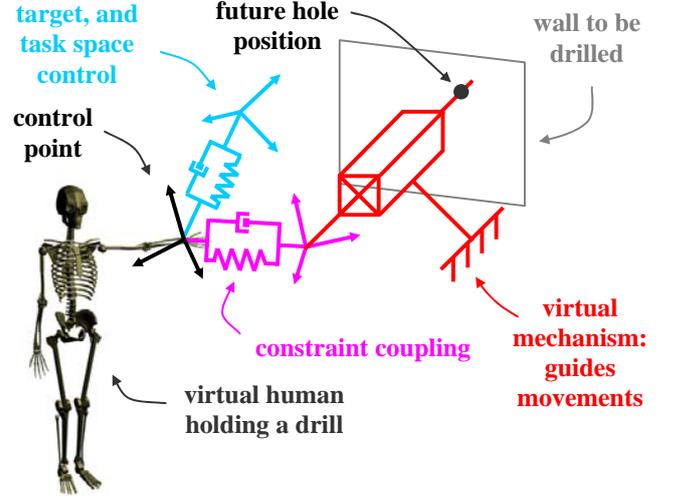

Fig. 5 Passively guided virtual human

In Fig. 5, the manikin holds a drill which axis is aligned with the future hole axis in the wall thanks to the simple virtual mechanism in red. Any other constraint could be expressed thanks to virtual mechanisms.

### IV. EXPERIMENTS AND RESULTS

In this section we test the virtual guides approach we have implemented. The experiment consists in drilling a hole in a wall thanks to a drill, while lighting the future hole location thanks to a hand light. Both tools are guided, thanks to our virtual mechanisms framework. The drill can only move along a fixed axis with a fixed orientation. This means that the controller leaves only one degree of freedom to the operator. The direction of the spotlight is also driven automatically (leaving the three degrees of freedom of the light's position to the operator). Fig. 6 depicts the ideal axis in green (a) and actual axis are in red (b).

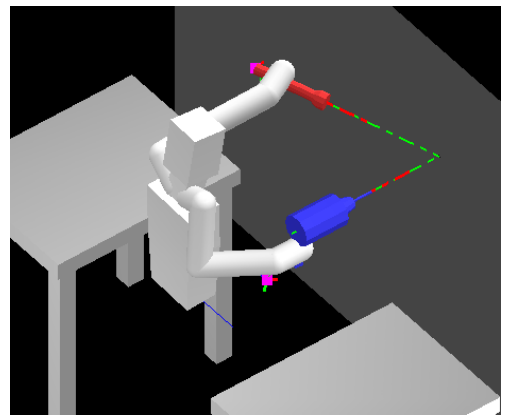

Fig. 6 Worker drilling a hole, guided by virtual mechanisms

In order to see the efficiency of our method, we drew the angle between the ideal axis, and the actual axis of the drill, as seen on Fig. 7; in the case where the operator is completely free (green), and in case where the guide is on (orange).

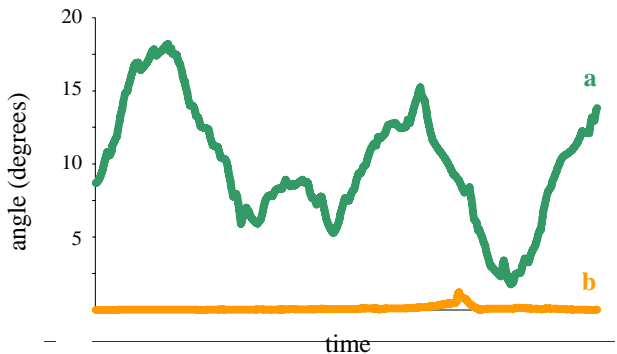

Fig. 7 Angle between ideal and actual axis of the drill, (a) without guide, and (b) with guide.

We also tested the collision engine. Fig. 8 shows anti-collision in action.

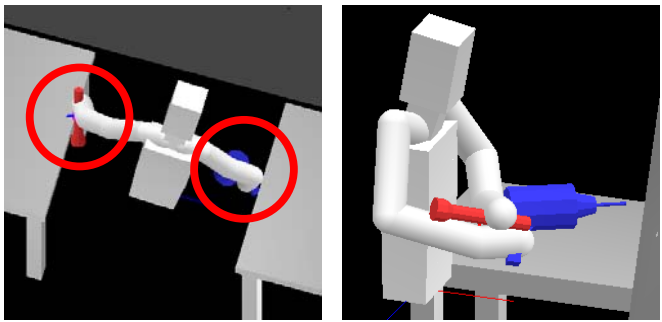

Fig. 8 Double, and self-collision

On the curve Fig. 9, we can see the height of the table, which must not be penetrated (orange), and the height of the virtual human's hand (green), while reaching, and leaning on the table. We see that the hand never penetrates the table.

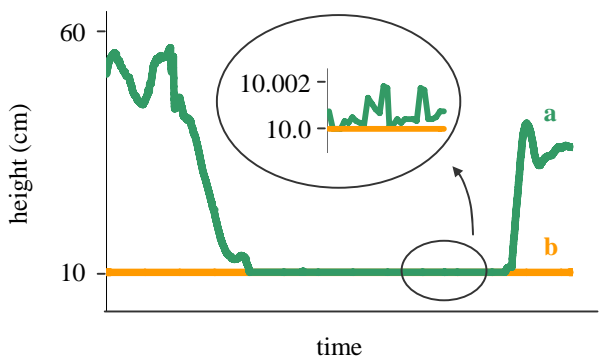

Fig. 9 Hand (a), and obstacle's (b) height: no penetration.

## V. CONCLUSION

We have introduced a new control scheme dedicated to controlling virtual humans, it decouples functionalities, and assure stability thanks to passivity.

This architecture makes it able to deal with task space, and internal control, unilateral constraints such as collision, and joint limits, and thanks to passivity, can be easily further extended. Nevertheless, we showed that projections could break passivity, hence they must be used carefully.

We have introduced a new tool as for manikin's control, which enables (i) to build passive projections, and (ii) to guide or help the virtual manikin achieve its task. This functionality was easily added to our controller thanks to its passive nature.

Next step will be to find how to build passive external tasks' projections.